\documentclass[journal]{IEEEtran}
\usepackage{amsmath,amsfonts}
\usepackage{algorithmic}
\usepackage{algorithm}
\usepackage{array}
\usepackage[caption=false,font=normalsize,labelfont=sf,textfont=sf]{subfig}
\usepackage{textcomp}
\usepackage{float}
\usepackage{stfloats}
\usepackage{url}
\usepackage{soul}
\usepackage{verbatim}
\usepackage{graphicx}
\usepackage{cite}
\usepackage{tikz}
\usepackage{bm}
\usepackage{fontawesome}
\usetikzlibrary{positioning}
\usepackage{overpic}
\usepackage{comment}
\usepackage{amsmath,amssymb} 
\usepackage{color}
\usepackage{animate}
\usepackage{makecell}
\usepackage{multirow}
\usepackage{booktabs}
\usepackage{sidecap}
\usepackage{import}

\usepackage[citecolor=blue, colorlinks]{hyperref}

\usepackage[numbers]{natbib}
\setstcolor{blue}
\begin{document}

\title{Learning A  Locally Unified 3D Point Cloud \\ for View Synthesis}

\author{Meng You, Mantang Guo, Xianqiang Lyu, Hui Liu, and Junhui Hou,~\IEEEmembership{Senior Member,~IEEE}
\thanks{This work was supported in part by the Hong Kong Research Grants
Council under Grants 11218121 and 21211518, in part by the Hong Kong
Innovation and Technology Fund under Grant MHP/117/21, in part by the
Basic Research General Program of Shenzhen Municipality under Grant
JCYJ20190808183003968, and in part by Hong Kong University Grants
Committee under Grant UGC/FDS11/E02/22. (Corresponding author: Junhui
Hou).}
\thanks{M. You, M. Guo, X. Lyu, and J. Hou are with the Department of Computer Science,
City University of Hong Kong, Hong Kong, and also with the City
University of Hong Kong Shenzhen Research Institute, Shenzhen 518057,
China. (e-mail:jh.hou@cityu.edu.hk)}
\thanks{H. Liu is with the School of Computing Information Sciences, Caritas Institute
of Higher Education, Hong Kong. (e-mail:hliu99-c@my.cityu.edu.hk)}}

\markboth{}%
{Shell \MakeLowercase{\textit{et al.}}: A Sample Article Using IEEEtran.cls for IEEE Journals}


\maketitle

\begin{abstract}
In this paper, we explore the problem of 3D point cloud representation-based view synthesis from a set of sparse source views.
To tackle this challenging problem, we propose a new deep learning-based view synthesis paradigm that learns a locally unified 3D point cloud from
source views. Specifically, we first construct sub-point clouds by projecting source views to 3D space based on their depth maps. Then, we learn the locally unified 3D point cloud by adaptively fusing points at a local neighborhood defined on the union of the sub-point clouds. 
Besides, we also propose a 3D geometry-guided image restoration module to fill the holes and recover high-frequency details of the rendered novel views. 
Experimental results on three benchmark datasets demonstrate that our method can improve the average PSNR by more than 4 dB while preserving more accurate visual details, compared with state-of-the-art view synthesis methods.
The code will be publicly available at \url{https://github.com/mengyou2/PCVS}.
\end{abstract}


\section{Introduction}
Given a collection of posed images observed from source views, view synthesis aims at generating photorealistic images at novel views.  As view synthesis can benefit a variety of applications, e.g., robotics \cite{manuelli2019kpam}, 3D modeling \cite{anderson2016jump,collet2015high}, virtual reality \cite{wei2019vr}, and so on, a considerable number of view synthesis methods \cite{zhou2016view,park2017transformation,sun2018multi,chen2019monocular,mildenhall2020nerf,shi2021self} have been proposed over the past decades. Particularly, 3D point cloud representation-based methods, 
which generally render novel views from 3D scene representations, e.g., 
3D point clouds/meshes, have been attracting attention.

\begin{figure*}[t]
\centering
\includegraphics[width=1\textwidth]{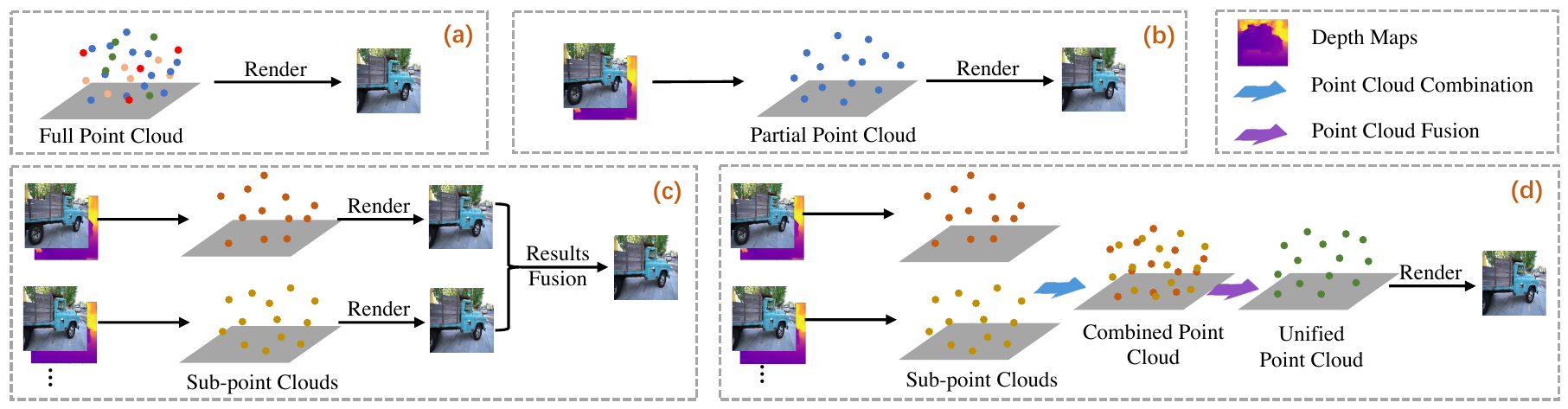} \vspace{-0.4cm}
\caption{
Comparison of different point cloud representation-based view synthesis paradigms. 
}
\label{figure:cmp}
\end{figure*}

Fig.~\ref{figure:cmp} shows paradigms of different 3D point cloud representation-based view synthesis methods. Specifically, 
some methods \cite{meshry2019neural,pittaluga2019revealing,aliev2020neural,dai2020neural,song2020deep,ruckert2022adop} illustrated in Fig.~\ref{figure:cmp} (\textcolor{red}{a}) require inputting the full 3D point cloud, which may be difficult and labor-intensive to obtain.
Conversely, 
other methods \cite{niklaus20193d,wiles2020synsin,le2020novel} shown in Fig.~\ref{figure:cmp} (\textcolor{red}{b}) construct the point cloud from \textit{a single} source view by predicting its depth map and further projecting the source view to 3D space. 
In reality, it is easy to obtain multiple source views of a scene, which may greatly improve the quality of the novel view synthesis. Unfortunately, it is not straightforward to generalize the single view-based paradigm shown in Fig.~\ref{figure:cmp} (\textcolor{red}{b}) to handle multiple input source views well. 
To synthesize novel views from more than one  source view \cite{cao2022fwd}, as shown in Fig.~\ref{figure:cmp} (\textcolor{red}{c}), one can simply extend the single view-based paradigm,  
i.e., separately generating an intermediate novel view from each source view and then fusing the intermediate results via a 
confidence-based blending module. 
However, such a manner cannot take advantage of the relationship between different source views well, thus limiting its performance (see Section \ref{sec:exp} for the results).

To this end,  we propose a new deep learning-based paradigm for view synthesis, as illustrated in Fig.~\ref{figure:cmp} (\textcolor{red}{d}), which learns a unified 3D point cloud by adaptively fusing the sub-point clouds constructed from different source views. Specifically, we first project pixels of each source view into 3D space with corresponding depth maps, leading to sub-point clouds. 
We consider two scenarios regarding depth maps: (1) datasets already provide them; (2) a typical self/un-supervised depth estimation method for multi-view images is utilized to estimate them.
As the depth maps are not perfect and inevitably contain errors, we adaptively fuse the sub-point clouds 
to a locally unified 3D point cloud representation with better quality via a point cloud fusion module. Besides, we propose a 3D geometry-guided image restoration module to fill holes and recover high-frequency details in the rendered novel view.
Extensive experiments on three benchmark datasets demonstrate the significant superiority of our method over state-of-the-art view synthesis methods both quantitatively and visually. 
Besides, comprehensive ablation studies validate the effectiveness of the key modules of our framework.


In summary, the main contributions of this paper are three-fold: 
\begin{itemize}
    \item  
    a new learning-based paradigm of point cloud representation-based view synthesis from multiple source views;
   \item  
   a learnable point cloud fusion module 
   to 
   construct a locally unified point cloud representation of the input scene; 
   and
   \item 
   a 3D geometry-guided image restoration module to fill holes and  recover high-frequency details of the rendered novel view.

\end{itemize}

The rest of this paper is organized as follows. Section \ref{sec:RW} briefly reviews related works. Section \ref{sec:proposed} presents the proposed framework for view synthesis, followed by comprehensive experiments and analyses in Section  \ref{sec:exp}.
Finally, Section  \ref{sec:exp} concludes the paper and discusses some future directions for improving the proposed view synthesis paradigm.

\section{Related Work}
\label{sec:RW}

View synthesis is a long-standing problem in computer vision/graphics. Traditional methods  \cite{buehler2001unstructured,debevec1996modeling,gortler1996lumigraph,levoy1996light,seitz1996view} implement image-based rendering with the idea of blending synthesized images from source views. However, they are usually time-consuming and require dense inputs to achieve high-quality results.
Recently, powerful deep learning has been widely used in the view synthesis area. Some methods \cite{zhou2016view,chen2019monocular,sun2018multi,park2017transformation} perform pixel interpolation in the novel view by employing a neural network to estimate appearance flow between viewpoints. These methods lack scene geometry in their models, thus limiting their performance, especially for real scenes. To overcome these limitations, some methods attempt to explicitly utilize scene geometry by learning scene representations from large datasets. These methods first reconstruct 3D geometry, such as volumetric representations, point cloud representations, and neural representations, from images, and then render novel views. 

\subsection{Volumetric Representation-based Methods}
Recently, some methods have been learning the volumetric representations, such as voxel-based grid, multi-plane image (MPI), or layered depth image (LDI), 
from source images. Specifically, 
the voxel-based grid methods \cite{yan2016perspective,choy20163d,jimenez2016unsupervised,kar2017learning,penner2017soft,xie2019pix2vox,henzler2019escaping,lombardi2019neural} represent objects as a 3D volume in the form of voxel occupancies. Kar \textit{et al.} \cite{kar2017learning} utilized the underlying 3D geometry in multi-view images to reconstruct the voxel occupancy grid by unprojecting image features along with viewing rays. Sitzmann \textit{et al.} \cite{sitzmann2019deepvoxels} proposed DeepVoxels to encode the view-dependent appearance of the scene without explicitly modeling its 3D geometry.
MPI \cite{zhou2018stereo,flynn2019deepview,mildenhall2019local,srinivasan2019pushing,li2020crowdsampling} represents the scene as a set of fronto-parallel planes at fixed depths, where each plane consists of an RGB image and an $\alpha$ map. Mildenhall \textit{et al.} \cite{mildenhall2019local} proposed to expand each input view into an MPI, and then render the novel view by blending its adjacent MPIs.
Li \textit{et al.} \cite{li2020crowdsampling} built the DeepMPI representation by adding latent features in MPI layers to render view-dependent lighting effects. 
Similar to MPI, LDI \cite{shade1998layered,swirski2011layered,tulsiani2018layer,shih20203d,dhamo2019object} keeps several depth and color values at every pixel and renders images by a back-to-front forward warping algorithm. Tulsiani \textit{et al.} \cite{tulsiani2018layer} used convolutional neural networks (CNN) to infer LDI representation from a single image and forward-splitting pixels to render a novel view. Shih \textit{et al.} \cite{shih20203d} proposed to generate LDI from RGBD images and employ a learning-based inpainting model to synthesize the color and depth information at occluded regions. 
Choi \textit{et al.} \cite{choi2019extreme} and Shi \textit{et al.} \cite{shi2021self} constructed the depth probability volume of the novel view to backward warp color images or feature maps from source views.
Because of their discrete sampling, these mentioned volumetric representation-based methods could not achieve high-resolution results.  To overcome this issue, we opt to use point clouds as the representation of the scene, which has the advantage of accurately representing complex and irregular geometries. Point clouds consist of individual points that can be positioned anywhere in 3D space, enabling them to capture intricate details and subtle features that might be challenging to depict with a voxel grid.

\subsection{3D Point Cloud Representation-based Methods}
Given the point cloud associated with descriptors, some methods \cite{meshry2019neural,pittaluga2019revealing,aliev2020neural,ruckert2022adop,rakhimov2022npbg++} rasterize the point cloud into 2D image space with a learning-based differentiable rendering scheme. Instead of explicitly rendering the point cloud representation, Dai \textit{et al.}  \cite{dai2020neural} and Song \textit{et al.} \cite{song2020deep} proposed to extract features from the point cloud representation to construct a multi-plane 3D representation, and then render the color image from it via a neural network. Other methods \cite{Riegler2020FVS,Riegler2021SVS} use the point cloud as a base geometry model, which is further fitted to a surface mesh, and generate novel views by blending weights of sources on the mesh surface. 
Some methods construct the point cloud by estimating the depth maps of source views.
Niklaus \textit{et al.} \cite{niklaus20193d} proposed to predict the depth map of the source view guided by semantic information and then render the novel view from the colored point cloud constructed from the source view based on the estimated depth map. 
Based on the estimated depth map, Wiles \textit{et al.} \cite{wiles2020synsin} projected the feature map of the source view to the 3D space and then synthesized a novel view by decoding the feature map rendered from the point cloud. Le \textit{et al.} \cite{le2020novel} proposed to backward warp the synthesized novel view to the source one to supervise the depth estimation of the source view.  Cao \textit{et al.} \cite{cao2022fwd} forward warped each input view with a differentiable point cloud renderer similar to \cite{wiles2020synsin}, but extended to multiple inputs by fusing rendered view-dependent features. 
 Alieve \textit{et al.} \cite{aliev2020neural} and Rakhimov \textit{et al.} \cite{rakhimov2022npbg++} also considered the problem of rendering views from point clouds associated with feature descriptors. 
For a specific scene, Alieve \textit{et al.} \cite{aliev2020neural} and Rakhimov \textit{et al.} \cite{rakhimov2022npbg++} 
constructed a complete 3D point cloud of the scene, derived from a large number of source images using SfM or MVS techniques, which is then utilized to render all target views.
During each training iteration, they only updated descriptors of points that were visible in the target view. Instead of reconstructing the entire scene,  we focus on locally learning partial point clouds guided by the target view,
i.e., given a set of source views, we obtain a point cloud from each source view and then fuse the resulting point clouds to produce a unified one that is further used to render in-between novel views. The unified point cloud varies with the set of input source views. 
Besides,  Alieve \textit{et al.} \cite{aliev2020neural} started with zero descriptor values and updated the neural descriptors via backpropagation through the loss derivatives. Rakhimov \textit{et al.} \cite{rakhimov2022npbg++} modeled the neural descriptor for the point as a linear combination of learnable basis functions, where basis functions were learned by MLPs with inputting view directions, and coefficients were obtained by solving multivariate linear regression problems. Our method constructs the neural descriptor for the point by concatenating the RGB value and image features
and generates the neural descriptor of the point in the unified point cloud by interpolating the neural descriptors of its $K$NN points. 


\subsection{Neural Scene Representation-based Methods}
More recent methods represent the scene as neural radiance field (NeRF) by learning a continuous volumetric scene function \cite{mildenhall2020nerf}.
Martin \textit{et al.} \cite{martin2021nerf} extended NeRF to handle a collection of in-the-wild images.
To tackle the problem that NeRF requires to be re-trained before generalizing to other unobserved scenes, 
some methods \cite{yu2021pixelnerf,Wang_2021_CVPR,trevithick2020grf,SRF,li2021mine,chen2021mvsnerf} focus on NeRF generalization by involving the scene prior at the training phase.
Yu \textit{et al.} \cite{yu2021pixelnerf} attached image features behind the inputs of the NeRF. Wang \textit{et al.} \cite{Wang_2021_CVPR} combined image-based rendering method with NeRF by aggregating input image features to estimate visibility and blend colors simultaneously. Chibane \textit{et al.} \cite{SRF} introduced the classical multi-view stereo ideas into NeRF, predicting RGB and density for each 3D point from its stereo correspondence in the image feature space. Chen \textit{et al.} \cite{chen2021mvsnerf} leveraged 3D plane-swept cost volumes to reconstruct a neural encoding volume with per-voxel neural features, which is further regressed to volume density and radiance. Xu \textit{et al.} \cite{xu2022point} introduced scene geometry into NeRF using 3D point clouds with neural descriptors to model a radiance field. 

 Compared with our method, NeRF-based methods generally require a large number of training images to learn a high-quality radiance field. 
As NeRF-based methods only supervise the final integral radiance color instead of directly supervising RGB colors and density values of those 3D points, it is highly ill-posed for predicting RGB colors and density values of sampled 3D points by a simple MLP with few training images.
Our method can generate high-quality target views with only a few source views, e.g., two source views. Thus, our method would be more suitable than NeRF-based methods for the scenario where a limited number of source views are available. 
Since NeRF-based methods model a scene as a continuous function in 3D space, they can produce smooth animations or transitions between synthesized views. For our method, we construct different unified 3D point clouds from different sets of input source views, which leads to slight inconsistency between synthesized views. Thus, NeRF-based methods would be more suitable than our method for the scenario requiring highly smooth animations between synthesized views.

\section{Proposed Method}
\label{sec:proposed}
\begin{figure*}[t]
\centering
\includegraphics[width=1\textwidth]{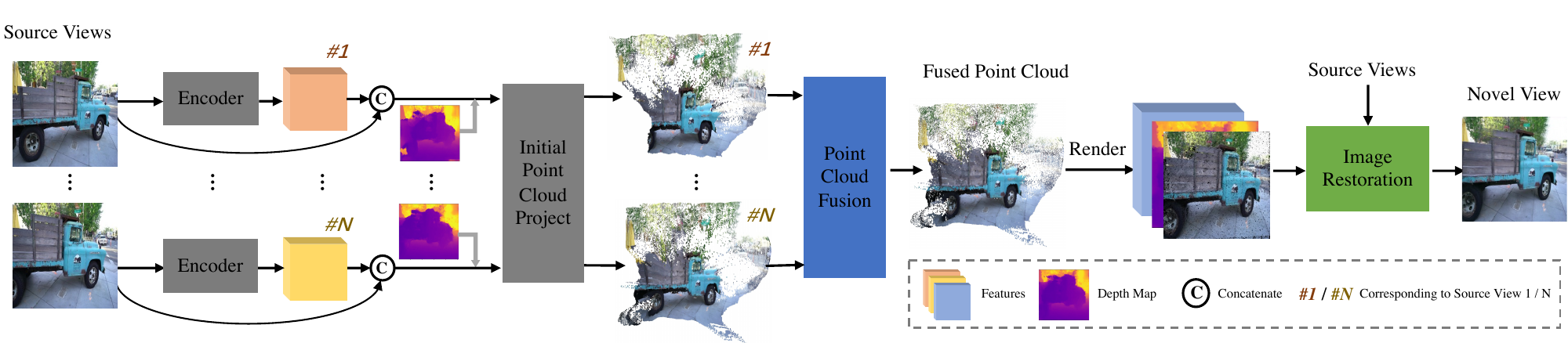}\vspace{-0.3cm}
\caption{
Flowchart of the proposed framework for view synthesis from multiple source views. 
It is mainly composed of two parts, i.e., learning of the locally unified point cloud from multiple sub-point clouds of input views and 3D geometry-guided image restoration, which are detailed in Fig. \ref{figure:pointfusion} and Fig. \ref{figure:restoration}, respectively. 
}
\label{figure: pipeline}
\end{figure*}
\textbf{Overview}. 
As shown in Fig.~\ref{figure: pipeline}, we consider synthesizing novel views  
from multiple source views by reconstructing a unified 3D point cloud as the scene representation.
Specifically, we first construct a sub-point cloud from each source view by projecting its pixels to 3D space based on the depth map and then fuse the sub-point clouds to build a locally unified one via the point cloud fusion module (Section \ref{sec: pointcloud}). After rendering the novel view from the unified point cloud, we restore it by filling holes and recovering high-frequency details 
via a 3D geometry-guided restoration module (Section \ref{sec: restoration}). In what follows, we detail each module.  

Note that rather than constructing a single 3D point cloud for a specific scene, which is then used for rendering all novel views, we aim to construct different locally unified 3D point clouds from different sets of source views, i.e., given a set of source views, we obtain a sub-point cloud from each source view and then fuse the resulting sub-point clouds to produce a locally unified one that is further used to render in-between novel views. The locally unified point cloud varies with the set of source views. Such a manner is potential memory- and computationally-efficient because a source view far away from the target view has a marginal contribution.

\subsection{Unified 3D Point Cloud Representation}
\label{sec: pointcloud}

Given $N$ source views of dimensions $H\times W$ $\left\{\mathbf{I}_{n}\in\mathbb{R}^{W\times H}\right\}_{n=1}^N$ and their depth maps $\left\{\mathbf{D}_{n}\in\mathbb{R}^{W\times H}\right\}_{n=1}^N$, as well as the camera intrinsic and extrinsic parameters, we can project the pixels of $\left\{\mathbf{I}_{n}\right\}_{n=1}^N$ to a common 3D coordinate system, producing $N$ sub-colored point clouds.
Ideally, pixels of $\left\{\mathbf{I}_{n}\right\}_{n=1}^N$ corresponding to the same scene point should be projected to an identical 3D point. 

\begin{figure*}[t]
\centering
\includegraphics[width=1\textwidth]{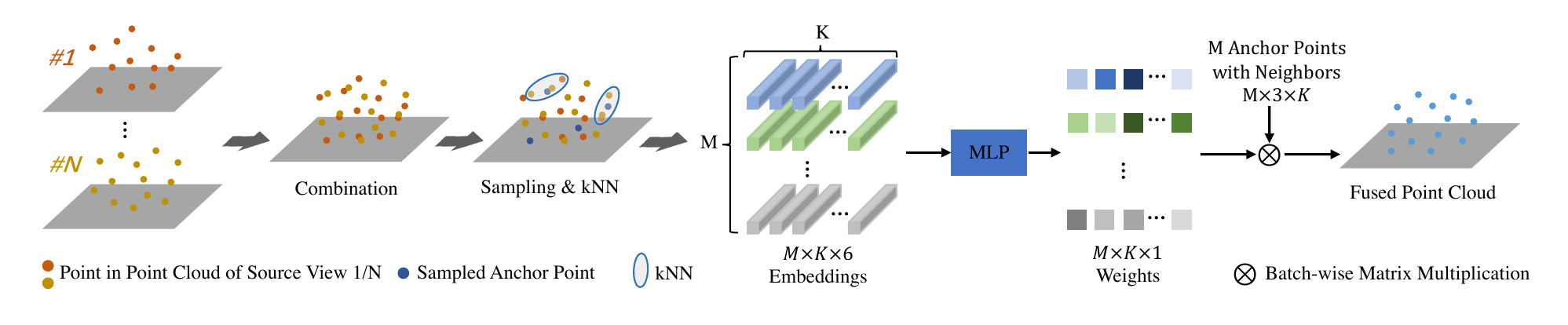}\vspace{-0.3cm}
\caption{Flowchart of the point cloud fusion module.
} 
\label{figure:pointfusion}
\end{figure*}

However, due to inevitable errors in estimated depth maps or occlusions among different source views, there would be point deviations, i.e., the pixels corresponding to the same scene point are projected to different 3D locations in the union of projected sub-point clouds. Thus, we propose a point cloud fusion module to fuse these sub-point clouds into a locally unified one for synthesizing novel views in-between source viewpoints. As shown in Fig.~\ref{figure:pointfusion}, we first sample some anchor points from the union of sub-point clouds randomly. For each anchor point, we then seek its K nearest neighboring ($K$NN) points, which are further linearly interpolated with learned weights via an MLP to synthesize the point of the unified point cloud. In this way, we can eliminate point deviations effectively by adaptively adjusting the point cloud, and significantly improve the reconstruction quality of novel views.

Specifically, let $\left\{\mathcal{P}_{n} \in \mathbb{R}^{S\times 3}\right\}_{n=1}^N$
denote the projected sub-point clouds corresponding to $\left\{\mathbf{I}_{n}\right\}_{n=1}^N$,
where $S=H\times W$. We first sample $M$ points named anchor points from the union of $\left\{\mathcal{P}_{n} \in \mathbb{R}^{S\times 3}\right\}_{n=1}^N$ in a random manner
to construct a base point cloud $\mathcal{P}_b=\left\{\mathbf{x}_i\in \mathbb{R}^3\right\}_{i=1}^M$, i.e.,
\begin{equation}
   \mathcal{P}_b= f_s(\mathcal{P}_{1}\cup \cdots  \mathcal{P}_{n}  \cdots \cup \mathcal{P}_{N}),  
\end{equation}
where 
$f_s(\cdot)$ denotes the random sampling process. 

For a typical anchor point $\mathbf{x}_i \subset\mathcal{P}_b$, we seek its $K$NN from $\mathcal{P}_{1}\cup \cdots \cup \mathcal{P}_{N}$ in the sense of Euclidean distance, denoted by $\left\{\mathbf{x}_i^k\in \mathbb{R}^3\right\}_{k=1}^K$. We then employ an MLP to learn the interpolation weight for a typical neighbor $\mathbf{x}_i^k$ by embedding the following information.\\

\noindent\textbf{(1) The relative position and distance between $\mathbf{x}_i^k$ and $\mathbf{x}_i$.} We define the relative position $\Delta \mathbf{x}_i^k$ and distance $d_i^k$ as the difference and the Euclidean distance between $\mathbf{x}_i^k$ and $\mathbf{x}_i$, respectively, i.e., 
\begin{equation}
  \Delta \mathbf{x}_i^k = \mathbf{x}_i^k - \mathbf{x}_i, ~{\rm and }~ d_i^k=\parallel \mathbf{x}_i^k - \mathbf{x}_i\parallel_2, 
\end{equation}
where $\parallel\cdot \parallel_2$ denotes the $\ell_2$ norm of a vector.\\


\noindent\textbf{(2) The descriptor similarity between $\mathbf{x}_i^k$ and $\mathbf{x}_i$.} We first separately construct a descriptor for $\mathbf{x}_i^k$ and $\mathbf{x}_i$, i.e.,  
\begin{equation}
  \widehat{\mathbf{f}}_i^k=\texttt{CAT}(\mathbf{c}_i^k, \mathbf{f}_i^k), ~{\rm and }~ \widehat{\mathbf{f}}_i=\texttt{CAT}(\mathbf{c}_i, \mathbf{f}_i), 
\end{equation}
where $\texttt{CAT}(\cdot)$ is the concatenation operation, and $\widehat{\mathbf{f}}_i^k \in \mathbb{R}^{35}$, $\mathbf{c}_i^k \in \mathbb{R}^3$ and $\mathbf{f}_i^k \in \mathbb{R}^{32}$ (resp. $\widehat{\mathbf{f}}_i \in \mathbb{R}^{35}$, $\mathbf{c}_i \in \mathbb{R}^3$ and $\mathbf{f}_i \in \mathbb{R}^{32}$) are the descriptor, the RGB values and the image feature corresponding to $\mathbf{x}_i^k$ (resp. $\mathbf{x}_i$), respectively, 
 and we learn the image features $\mathbf{f}_i^k$ and $\mathbf{f}_i$ from source views by employing a sub-CNN. Then, we compute the descriptor similarity $s_i^k$ as the cosine similarity between $\widehat{\mathbf{f}}_i^k$ and $\widehat{\mathbf{f}}_i$, i.e.,
\begin{equation}
  s_i^k = \frac{\widehat{\mathbf{f}}_i^k\cdot \widehat{\mathbf{f}}_i}{\mid\widehat{\mathbf{f}}_i^k\mid \mid\widehat{\mathbf{f}}_i\mid}.
\end{equation}
We finally construct the embedding $\mathbf{e}_i^k$ as 
\begin{equation}
\label{equ:embedding}
  \mathbf{e}_i^k=\texttt{CAT}(\Delta \mathbf{x}_i^k, d_i^k, s_i^k),  
\end{equation}
and separately predict the interpolation weight $w_p^{i,k}$ and $w_f^{i,k}$ for the point position and its corresponding descriptor as 
\begin{equation}
  w_p^{i,k}=f_p(\mathbf{e}_i^k;\bm{\theta}_p), ~{\rm and }~~ 
  w_f^{i,k}=f_c(\mathbf{e}_i^k;\bm{\theta}_c), 
\end{equation}
where $f_p(\cdot;\cdot)$ and $f_c(\cdot;\cdot)$ are the learnable MLPs parameterized by $\bm{\theta}_p$ and $\bm{\theta}_c$, respectively. With the learned weights, we can obtain the point and its corresponding descriptor of the unified 3D point cloud $\widetilde{\mathcal{P}}=\{\widetilde{\mathbf{x}}_i\in\mathbb{R}^3\}_{i=1}^M$ as 
 \begin{equation}
  \widetilde{\mathbf{x}}_i=\sum_{k=1}^K w_p^{i,k} \mathbf{x}_i^k, ~{\rm and }~ 
  \widetilde{\mathbf{f}}_i=\sum_{k=1}^K w_f^{i,k} \widehat{\mathbf{f}}_i^k.
\end{equation}

\textit{Remark.} Random sampling has demonstrated effectiveness in deep learning-based semantic segmentation for large-scale point clouds \cite{hu2020randla}. When the sampling rate is high, random sampling can efficiently produce an approximate uniformly-distributed sub-set. 
Besides, another potential advantage of random sampling is that it can introduce greater diversity in the training data, making our method more robust and generalizing better to new and unseen point clouds. 
Since the sampled anchor point cloud is not directly used for rendering but rather to find subsets of the union of sub-point clouds by searching $K$NN, any information loss resulting from random sampling can be compensated by fusing all points in the subset to gather local information and encode it into the interpolated descriptor. 
We also refer readers to the experimental validation of the effectiveness and advantages of random sampling for this task in Section \ref{subsec:ablation}.

 \begin{figure*}[t]
\centering
\includegraphics[width=1\textwidth]{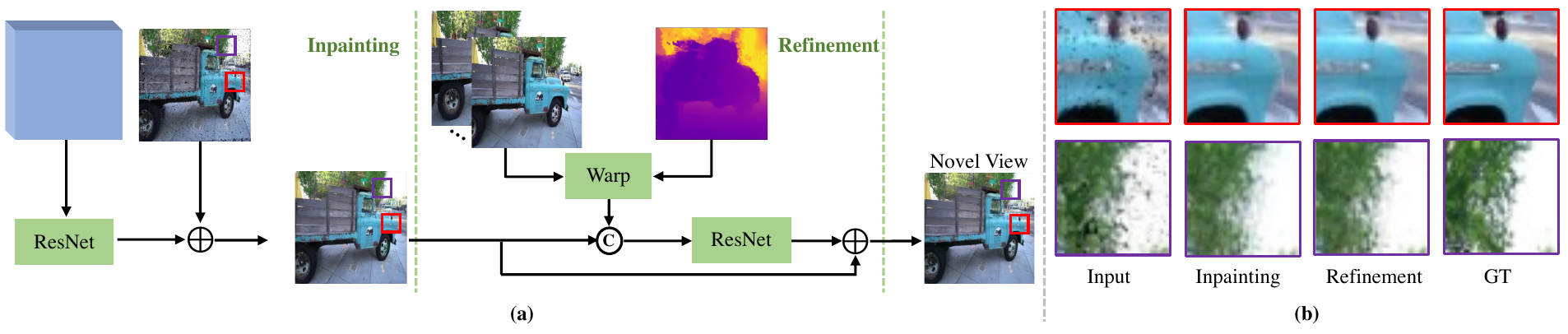}\vspace{-0.4cm}
\caption{(a) Flowchart of the 3D geometry-guided image restoration module. (b) Visual illustration of the effect of this module. 
}
\label{figure:restoration}
\end{figure*}

\subsection{3D Geometry-guided Image Restoration}
\label{sec: restoration}
With the unified 3D point cloud representation as well as the camera intrinsic and extrinsic parameters of the target view, we can render a coarse novel view image, denoted as $\widetilde{\mathbf{I}}_t^c$, via a typical renderer. In this paper, we adopt the differentiable renderer involved in PyTorch3D \cite{ravi2020pytorch3d} that splats each point to a circular region in screen-space whose opacity decreases away from the region's center and computes the value of each pixel by blending information for the neighboring points whose splatted regions overlap the pixel. However, as illustrated in the first column of Fig.~\ref{figure:restoration} (b), $\widetilde{\mathbf{I}}_t^c$ 
still suffers from 
holes since there are still pixel positions not in the splatted regions of any points. Besides, blending the colors of several points into one pixel also causes the missing of high-frequency details in $\widetilde{\mathbf{I}}_t^c$. 
To deal with these issues, as shown in Fig.~\ref{figure:restoration} (a), we propose a geometry-guided image restoration module to restore $\widetilde{\mathbf{I}}_t^c$ by filling holes and propagating high-frequency details from the source views under the guidance of the depth map of the novel view.

Specifically, we first project $\widetilde{\mathcal{P}}$ 
associated with descriptors to render a feature map of the novel view, denoted by $\widehat{\mathbf{F}}_t$. To fill the holes in $\widetilde{\mathbf{I}}_t^c$, we learn an additive map by using a sub-CNN $f_h(\cdot;\cdot)$ parameterized by $\bm{\theta}_h$ with $\widehat{\mathbf{F}}_t$ as input, i.e., 
 \begin{equation}
    \widetilde{\mathbf{I}}_t^p= f_h(\widehat{\mathbf{F}}_t;\bm{\theta}_h)+\widetilde{\mathbf{I}}_t^c.
 \end{equation}
Furthermore, we transform $\widetilde{\mathcal{P}}$
to the camera coordinate system of the novel view, and then project the points associated with $z$-coordinates to render the depth map of the novel view, based on which, we separately backward warp each source view in $\left\{\mathbf{I}_{n}\right\}_{n=1}^N$ to the novel view, generating warped source images $\left\{\widehat{\mathbf{I}}_{n}\right\}_{n=1}^N$
. To propagate the high-frequency details from the source views to $\widetilde{\mathbf{I}}_t^p$, we also learn an additive map by using another sub-CNN $f_r(\cdot;\cdot)$ parameterized by $\bm{\theta}_r$ with the concatenation of $\widetilde{\mathbf{I}}_t^p$ and warped source images
as input, i.e., 
 \begin{equation}
 \label{eq:refine}
     \widetilde{\mathbf{I}}_t= f_r\left(\texttt{CAT}(\widetilde{\mathbf{I}}_t^p,\widehat{\mathbf{I}}_{1},\cdots,\widehat{\mathbf{I}}_{n},\cdots, \widehat{\mathbf{I}}_{N});\bm{\theta}_r   \right)+\widetilde{\mathbf{I}}_t^p,
 \end{equation}
where $\widetilde{\mathbf{I}}_t$ is the finally synthesized novel view. 
As an example, Fig.~\ref{figure:restoration} (\textcolor{red}{b}) visually illustrates the effect of the proposed 3D geometry-guided image restoration module.

\textit{Remark.} From a technical point of view, the order of warped images (i.e., the orders of warped images are inconsistent during training and inference) could affect the performance of our method because the warped source images could differ from each other to a large extent due to different relative positions between source views and the target view,  
resulting in different contributions to the synthesized view (see the experimental verification in Section \ref{subsubsec:order}).
Thus, to eliminate the order effect, we should keep the order of source views consistent during training and inference.

\subsection{Loss Function}
We supervise both the intermediate and final rendered results by calculating the photometric loss with the ground-truth novel view $\mathbf{I}_t$. We define the photometric loss function $\ell_r(\cdot,\cdot)$ as 
\begin{equation}
    \ell_r(\widetilde{\mathbf{I}}_t^{*},\mathbf{I}_t) = \left \|\widetilde{\mathbf{I}}_t^{*}-\mathbf{I}_t\right\|_1+\sum_{l=1}^{L} \lambda_l \left \|\phi_l(\widetilde{\mathbf{I}}_t^{*})-\phi_l(\mathbf{I}_t) \right\|_1, 
\end{equation}
where $\left\|\cdot\right\|_1$ denotes the $\ell_1$ norm, $\widetilde{\mathbf{I}}_t^{*} \in \{\widetilde{\mathbf{I}}_t^{c}, \widetilde{\mathbf{I}}_t^{p}, \widetilde{\mathbf{I}}_t\}$, $ \left\{\phi_l\right\}_{l=1}^{L}$ is a set of layers in a pre-trained VGG-19 network \cite{simonyan2014very}, the weights $ \left\{\lambda_l\right\}_{l=1}^{L}$ are set to the inverse of the number of neurons in each layer, and $L=5$. Thus, the total photometric loss is calculated as 
\begin{equation}
    \widehat{\ell}_r = \ell_r(\widetilde{\mathbf{I}}_t^{c},\mathbf{I}_t) + \ell_r(\widetilde{\mathbf{I}}_t^{p},\mathbf{I}_t) + \ell_r(\widetilde{\mathbf{I}}_t,\mathbf{I}_t).  
\end{equation}

\subsection{Practical Extension to the Scenario without Depth Maps}
As depth maps are not always available in practice, we also extend our method by plugging a self-supervised depth estimation module, which is jointly trained with our framework. Such an extension can demonstrate the generalization ability of our method. We employ the network of MVSNet \cite{yao2018mvsnet}, a multi-view-based depth estimation network, to separately estimate the depth maps of input source views. 
However, 
MVSNet \cite{yao2018mvsnet} requires ground-truth depth maps as supervision during training.
To this end, we adopt a self-supervised loss term to regularize the learning of this network. 

 Specifically, to regularize MVSNet \cite{yao2018mvsnet} generating the depth map of a typical source view, we warp other source views and the ground-truth target view to the current source view based on the estimated depth map, and then minimize the errors between the current source view and the warped images. 
Specifically, to estimate the depth map $\mathbf{D}_{n}$ ($n\in [1,N] $) of a typical source view $\mathbf{I}_{n}$,
we inversely warp 
other views $\left\{\mathbf{I}_{i}\right\}_{i=1,i\neq n}^N$ and the ground-truth novel view $\mathbf{I}_t$ to $\mathbf{I}_{n}$ based on $\mathbf{D}_{n}$, leading to the warped images $\left\{\mathbf{I}_{i\rightarrow n}\right\}_{i=1,i\neq n}^N$ and $\mathbf{I}_{t\rightarrow n}$, respectively. 
Formally, we write the self-supervised loss term for estimating $\mathbf{D}_{n}$ as  
\begin{equation}
\begin{aligned}
    \ell_{dn}^{self} = &\left \| (\mathbf{I}_{n} - \mathbf{I}_{t\rightarrow n})\odot \mathbf{M}_{t\rightarrow n} \right\|_1\\
    &+\sum_{i=1}^N \left \| (\mathbf{I}_{n} - \mathbf{I}_{i\rightarrow n})\odot \mathbf{M}_{i\rightarrow n} \right\|_1 ,i\neq n,
\end{aligned}
\label{equ:depth_self}
\end{equation}
where $\mathbf{M}_{i\rightarrow n}$ is the binary mask corresponding to $\mathbf{I}_{i\rightarrow n}$, where $0$ (resp. $1$) indicates the projected pixel is out of (resp. in) the range of the warped image. 
To further regularize the depth estimation module, apart from the $L1$ loss between the warped image and the source view, i.e., Eq.~\eqref{equ:depth_self}, we also add the SSIM loss between them. For a typical pair of warped image and source view, we define the SSIM loss function $ \ell_d^{ssim}(\cdot,\cdot, \cdot)$ as  
\begin{equation}
    \ell_d^{ssim}(\mathbf{X},\mathbf{Y}, \mathbf{M}) = \frac{1-SSIM(\mathbf{M}\odot\mathbf{X},\mathbf{M}\odot\mathbf{Y})}{2},
\end{equation}
where $\mathbf{X}$, $\mathbf{Y}$, $\mathbf{M}$ denote the warped image, the source view, and the corresponding mask, respectively. Thus, we calculate the SSIM loss $\widehat{\ell}_{dk}^{ssim}$ for estimating $\mathbf{D}_n$  as
\begin{equation}
\begin{aligned}
    \widehat{\ell}_{dn}^{ssim} = &\ell_d^{ssim}(\mathbf{I}_{n},\mathbf{I}_{t\rightarrow n},\mathbf{M}_{t\rightarrow n})\\
    &+\sum_{i=1}^N\ell_d^{ssim}(\mathbf{I}_{n},\mathbf{I}_{i\rightarrow n},\mathbf{M}_{i\rightarrow n}), i\neq n.
\end{aligned}
\end{equation}
Moreover, to promote smoothness of $\mathbf{D}_{n}$ , we penalize the $\ell_1$ norm of the gradient, denoted as $\ell_{dn}^{smooth}$: 
\begin{equation}
    \ell_{dn}^{smooth}=\left\|\nabla_x\mathbf{D}_{n}\right\|_1+\left\|\nabla_y\mathbf{D}_{n}\right\|_1,
\end{equation}
where $\nabla_x$ and $\nabla_y$ are the gradient operators for the spatial domain.
Thus, the total depth estimation loss is calculated as 
\begin{equation}
\begin{aligned}
    \ell_{d}=
    \sum_{n=1}^N(\lambda_{d1} \ell_{dn}^{self}
    +\lambda_{d2} \widehat{\ell}_{dn}^{ssim} 
    +\lambda_{d3} \ell_{dn}^{smooth}),
\end{aligned}
\end{equation}
where we empirically set $\lambda_{d1}=12$, $\lambda_{d2}=6$, and $\lambda_{d3}=0.18$.
If we employ the self-supervised depth estimation module, we calculate the total training loss as
\begin{equation}
\begin{aligned}
    \ell=\widehat{\ell}_r+\ell_d.
\end{aligned}
\end{equation}
Considering that estimated depth maps have deviations, we measure the depth estimation quality, namely probability map, proposed in MVSNet \cite{yao2018mvsnet}, which takes the probability sum over the four nearest depth hypotheses based on the estimated depth value to measure the estimation quality, as an added embedding term in the point cloud fusion module. We denote the depth estimation quality of $\mathbf{x}_i^k$ by $p_i^k$
and rewrite Eq.~\eqref{equ:embedding} as
\begin{equation}
  \mathbf{e}_i^k=\texttt{CAT}(\Delta \mathbf{x}_i^k, d_i^k, p_i^k, s_i^k).
\end{equation}

\begin{table*}[t]
\centering
\begin{center}
\caption{Quantitative comparison of different methods on \textit{Tanks and Temples}, \textit{DTU} and \textit{RealEstate10K}. We excluded the FVS method, Post-Fusion-D, and Ours-D for \textit{RealEstate10K} due to the lack of depth maps. $\uparrow$ (resp. $\downarrow$) means the larger (resp. smaller), the better. The best results are highlighted in bold. 
}
\label{tabel:all}
\begin{tabular}{cccc|ccc|ccc}
\toprule[1.2pt]
      & \multicolumn{3}{c|}{Tanks and Temples}  & \multicolumn{3}{c|}{DTU}    & \multicolumn{3}{c}{RealEstate10K}\\ 
      & \multicolumn{1}{c}{LPIPS$\downarrow$} & \multicolumn{1}{c}{PSNR$\uparrow$} & \multicolumn{1}{c|}{SSIM$\uparrow$} & \multicolumn{1}{c}{LPIPS$\downarrow$} & \multicolumn{1}{c}{PSNR$\uparrow$} & \multicolumn{1}{c|}{SSIM$\uparrow$} & \multicolumn{1}{c}{LPIPS$\downarrow$} & \multicolumn{1}{c}{PSNR$\uparrow$} & \multicolumn{1}{c}{SSIM$\uparrow$}  \\ \hline
IBRNet \cite{Wang_2021_CVPR} &0.321 &19.55 &0.649 &0.288 &19.94 &0.713 &0.080&29.81&0.919
\\
SVNVS \cite{shi2021self} & 0.218&20.80 &0.730 &0.541 &15.50 &0.394 & 0.116&27.11 & 0.897
\\
Ours-W &0.189 &21.28 & 0.751&0.140 & 24.24&0.859&\textbf{0.028}&\textbf{36.88}&\textbf{0.975}\\ \hline 
FVS \cite{Riegler2020FVS}  &0.302 &18.91 &0.661 &0.236 &20.14 &0.723 & -&- & - \\
Post-Fusion-D &0.155 &23.47 &0.848 &0.193 &22.50 &0.823 & -&-&- \\
Ours-D &\textbf{0.128} &\textbf{24.77} & \textbf{0.869}&\textbf{0.138} &\textbf{24.59}&\textbf{0.874} &-&-&-\\
\bottomrule[1.2pt]
\end{tabular}
\end{center}
\end{table*}

\section{Experiments}
\label{sec:exp}
\subsection{Experiment Settings}
\subsubsection{Datasets} 
\label{sec:dataset}
We conducted 
extensive experiments on three challenging datasets, including RealEstate10K \cite{zhou2018stereo}, Tanks and Temples \cite{knapitsch2017tanks}, and DTU \cite{aanaes2016large}.  All three datasets contain rotations and translations in camera movement, and particularly the first one has minor movement and the latter two are more significant. 

 Specifically, \textit{RealEstate10K} is a huge dataset derived from 80k video clips and we chose a subset of 85 scenes for training and 17 scenes for testing. \textit{Tanks and Temples} contains more complex indoor and outdoor scenes with irregular camera trajectories. Following the settings in \cite{Riegler2020FVS}, we used 17 of 21 scenes for training and the remaining four (i.e., Truck, Train, M60, and Playground) for testing. \cite{Riegler2020FVS} also provides  depth maps for \textit{Tanks and Temples}, which were derived from a 3D surface mesh reconstructed from all views.
 \textit{DTU} consists of 124 different scenes, where each was captured by 49 cameras located regularly on a sphere. We used \textit{DTU} only for testing to verify the generalization ability. Particularly, we adopted the evaluation scene set of 18 scenes provided by Yao \textit{et al.} \cite{yao2018mvsnet}. \cite{yao2018mvsnet} also provides depth maps for \textit{DTU} by rendering the 3D mesh reconstructed from the ground-truth point cloud to each viewpoint.
 
For RealEstate10K, Tanks and Temples, and DTU datasets, we tested all views of a typical test scene, i.e., all views in a test scene are test views. For each view of the test scene, we first designated it as the target view, and then selected a specific number (2, 3, or 4) of source views for it. 
 Specifically, for the 2-input setting, we choose the left and right adjacent frames of the target view as source views. For the setting with more than two source views, in addition to selecting two adjacent frames of the target view, we adopted the view selection result by \cite{Riegler2020FVS} to add more source views.

\subsubsection{Implementation details}
We implemented the encoder using Res-UNet \cite{xiao2018weighted} for image feature extraction and set the feature dimensions as 32.
For cost volume construction in our self-supervised depth estimation module, we sampled $D=128$ depth plane layers uniformly within the depth range of the scene. 
We implemented the inpainting and refinement components contained in the restoration module using ResNet with 6 blocks and 4 blocks, respectively. During training, 
we first trained the network without the refinement component $f_r(\cdot)$.
Then, we trained the refinement component $f_r(\cdot)$ with prior network parameters fixed. We used the Adam optimizer with the learning rate equal to $1e^{-5}$.

\begin{table*}[t]
\centering
\begin{center}

\caption{Quantitative comparison of different methods on the four scenes of \textit{Tanks and Temples}. The best results are highlighted in bold. 
} 
\label{tabel:tanks}\vspace{-0.3cm}
\begin{tabular}{cccc|ccc|ccc|ccc}
\toprule[1.2pt]
\label{table:headings}
    & \multicolumn{3}{c|}{Train}    & \multicolumn{3}{c|}{Playground}            & \multicolumn{3}{c|}{M60}     & \multicolumn{3}{c}{Truck}    \\ 
      & \multicolumn{1}{c}{LPIPS$\downarrow$} & \multicolumn{1}{c}{PSNR$\uparrow$} & \multicolumn{1}{c|}{SSIM$\uparrow$} & \multicolumn{1}{c}{LPIPS$\downarrow$} & \multicolumn{1}{c}{PSNR$\uparrow$} & \multicolumn{1}{c|}{SSIM$\uparrow$} & \multicolumn{1}{c}{LPIPS$\downarrow$} & \multicolumn{1}{c}{PSNR$\uparrow$} & \multicolumn{1}{c|}{SSIM$\uparrow$} & \multicolumn{1}{c}{LPIPS$\downarrow$} & \multicolumn{1}{c}{PSNR$\uparrow$} & \multicolumn{1}{c}{SSIM$\uparrow$} \\ \hline
IBRNet \cite{Wang_2021_CVPR} & 0.378& 17.37& 0.538&0.291 &23.01 &0.726 &0.287 & 18.82& 0.705 & 0.311& 19.92&0.660        
\\
SVNVS \cite{shi2021self}  &0.232 & 19.73& 0.674&0.204 &22.98 & 0.770& 0.210 & 20.29& 0.773 &0.222 & 20.58&0.715        
\\
Ours-W  &0.233 &20.04&	0.692 &0.158 &23.29 &0.780& 0.167 &21.10 & 0.808 &	0.183 &20.97 &  0.732  \\\hline
FVS \cite{Riegler2020FVS}  &0.341& 18.22& 0.597&0.286 &20.91 &0.696 &0.354 &16.28 &0.633  &0.180 &21.32 & 0.768\\
Post-Fusion-D &0.167&21.82&0.812&0.156&26.19&0.876&0.162&22.42&0.861&0.123&24.36&0.856\\
Ours-D &\textbf{0.145} & \textbf{22.63}&	\textbf{0.835} &\textbf{0.119} &\textbf{27.50} &\textbf{0.887} &\textbf{0.121}&\textbf{24.78} &\textbf{0.897} &	\textbf{0.118} &\textbf{24.92} &  \textbf{0.865}  \\
\bottomrule[1.2pt]
\end{tabular}
\end{center}
\end{table*}

\begin{figure}[h]
\centering
\vspace{-0.4cm}
\includegraphics[width=0.5\textwidth]{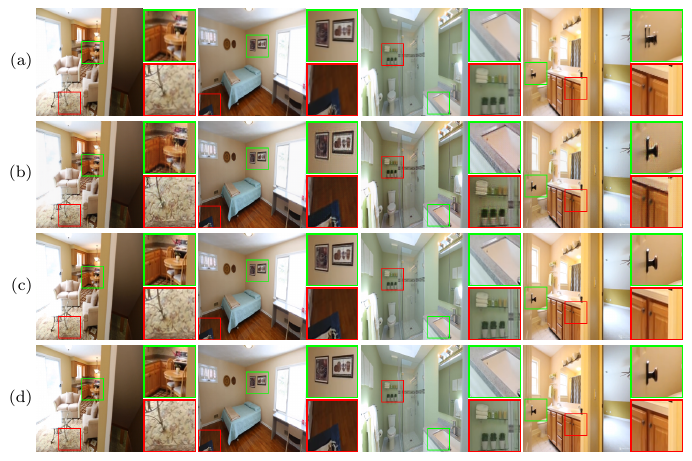} 
\vspace{-0.4cm}
\caption{
Visual comparison of different methods on \textit{RealEstate10K}. (a) IBRNet \cite{Wang_2021_CVPR}, (b) SVNVS \cite{shi2021self},  (c) Ours-W, (d) Ground Truth. \color{cyan}{\faSearch~} Zoom in to see details.}
\label{figure:re10k}
\end{figure}

\begin{figure}[h]
\centering
\includegraphics[width=0.5\textwidth]{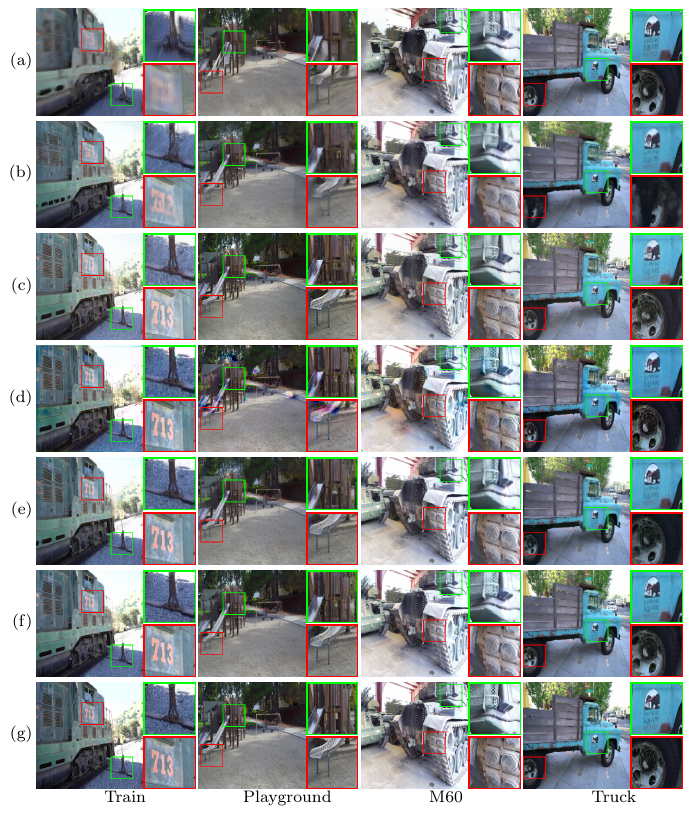} 
\vspace{-0.4cm}
\caption{Visual comparison of different methods on \textit{Tanks and Temples}. (a) FVS \cite{Riegler2020FVS}, (b) IBRNet \cite{Wang_2021_CVPR}, (c) SVNVS \cite{shi2021self}, (d) Post-Fusion-D, (e) Ours-W, (f) Ours-D, (g) Ground Truth. \color{cyan}{\faSearch~} Zoom in to see details.}
\label{figure:tanks}
\vspace{-0.4cm}
\end{figure}

\begin{figure}[h]
\centering
\includegraphics[width=0.5\textwidth]{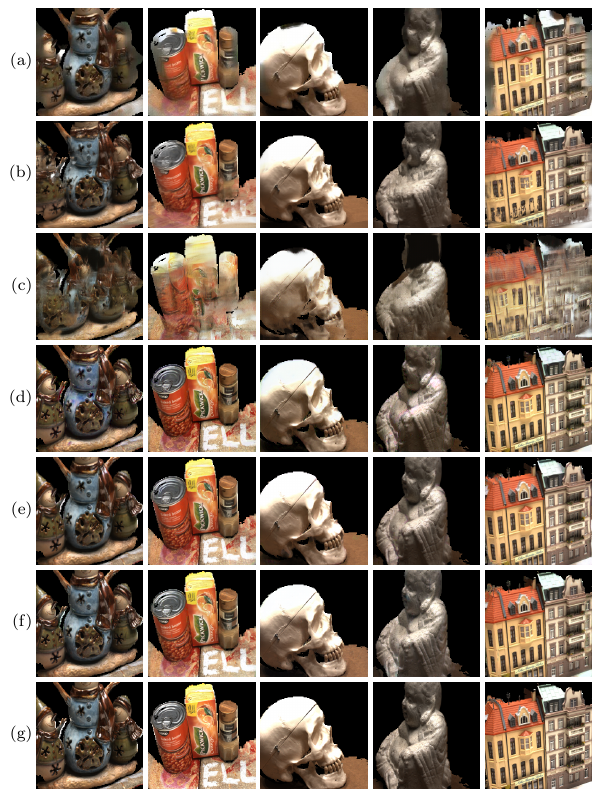} 
\vspace{-0.4cm}
\caption{Visual comparison of different methods on \textit{DTU}. (a) FVS \cite{Riegler2020FVS}, (b) IBRNet \cite{Wang_2021_CVPR}, (c) SVNVS \cite{shi2021self}, (d) Post-Fusion-D, (e) Ours-W, (f) Ours-D, (g) Ground Truth. \color{cyan}{\faSearch~} Zoom in to see details.}
\label{figure:dtu}
\vspace{-0.4cm}
\end{figure}


\subsection{Comparison with State-of-the-Art Methods}
We compared our method with the three most recent view synthesis methods under the setting of synthesizing novel views from two input source views, including two  image-based rendering methods, i.e., FVS \cite{Riegler2020FVS} and SVNVS \cite{shi2021self}, and one NeRF-based method, i.e., IBRNet \cite{Wang_2021_CVPR}. 
For fair comparisons, we retrained and evaluated these methods with the same data and settings as ours and well-tuned hyperparameters.
We named our method \textit{Ours-D} when inputting pre-processed depth maps and \textit{Ours-W} when using  self-supervised depth estimation.
To directly demonstrate the effectiveness of the proposed unified 3D point cloud representation, we also constructed a baseline shown in Fig.~\ref{figure:cmp}(c) called \textit{Post-Fusion-D} by replacing our point cloud fusion module with the confidence-based blending of rendered results. Specifically, we projected each source view with its features to 3D space and rendered it to the target viewpoint separately. The rendered image 
was inpainted with the corresponding rendered feature map. Then we added a CNN to learn confidence maps for these inpainted images and fused them into one result further refined by warped source views. The encoding, inpainting, and refinement networks are the same as our method.

\subsubsection{Quantitative comparisons} 
Table~\ref{tabel:all} lists the average PSNR, SSIM, and LPIPS \cite{zhang2018unreasonable} of different methods on each of the three datasets, 
Besides, Table~\ref{tabel:tanks} lists the results of the four scenes contained in \textit{Tanks and Temples}. 
From Tables~\ref{tabel:all} and \ref{tabel:tanks}, it can be observed that  
\begin{itemize}
    \item[$\bullet$] both \textit{Ours-W} and \textit{Ours-D} consistently outperform other methods on all datasets, including both minor and  significant movement, and especially \textit{Ours-W} improves the PSNR of the second best method by more than 7 dB on  \textit{RealEstate10K}, demonstrating its significant superiority;  
    \item[$\bullet$] we used the models trained on \textit{Tanks and Temples} to perform testing on  \textit{DTU}. 
    The superiority of \textit{Ours-W}  and \textit{Ours-D} over other compared methods on \textit{DTU} is more prominent than that on 
    \textit{Tanks and Temples}, demonstrating its stronger generalization ability; 
    \item[$\bullet$] 
    compared with the results on \textit{Tanks and Temples}, the performance of all methods improves significantly on \textit{RealEstate10K}.  
    The reason may be that the camera movement of \textit{RealEstate10K} is minor, conducive to obtaining more accurate geometry information; 
     \item[$\bullet$] \textit{Ours-D} improves the reconstruction quality by 1.3 dB and 2.0 dB on \textit{Tanks and Temples} and \textit{DTU}, respectively, compared with the  baseline \textit{Post-Fusion-D}, demonstrating the superiority of unifying sub-point clouds by our 3D point cloud fusion strategy over 2D confidence-based image blending; 
     
    \item[$\bullet$] \textit{Ours-D} performs better  than \textit{Ours-W} on \textit{Tanks and Temples} by more than 3dB. The reason is that self-supervised depth estimation can be challenging for datasets with large and irregular camera movements, especially when given only a few input images. 
\end{itemize}

\subsubsection{Visual comparisons} We also visually compared the results of different methods on  \textit{Tanks and Temples}, \textit{DTU}, and \textit{RealEstate10K} in Fig.~\ref{figure:tanks}, Fig.~\ref{figure:dtu}, and Fig.~\ref{figure:re10k}, respectively, where it can be observed that our method can produce better details than all the compared methods for all datasets under the same experimental configuration.
Using two source views and depth maps, the FVS can not build an accurate 3D mesh, resulting in noticeable blurred artifacts, especially on the boundaries of the synthesized image. For the IBRNet, the blurred artifacts and ghost effects appear at high-frequency regions and occlusion boundaries. As a NeRF-based method, IBRNet needs more source views to achieve satisfactory results. For SVNVS, there are blurred artifacts and slight color distortions compared with the ground truth. For the Post-Fusion-D, the edges of results show severe color distortions and artifacts. Additionally,  Fig.~\ref{figure:dtu} shows the generalization ability advantages of \textit{Ours-W} and \textit{Ours-D}, and both of them can get sharp and clear results.

\subsubsection{Efficiency comparisons} 
Table~\ref{tabel:inference} compares the inference time of different methods on  \textit{DTU}. Note that we implemented all methods on a Linux server with Intel CPU Xeon Gold 6226R @ 2.90GHz, 512GB RAM, and NVIDIA GeForce RTX 3090 GPU. \textit{Ours-W} takes more time to synthesize novel views than \textit{Ours-D} due to the self-supervised depth estimation module. \textit{Ours-D} is slightly slower than FVS but faster than SVNVS and IBRNet. Taking the reconstruction quality and efficiency together, we believe our method is the best.

\begin{table}[t]
\centering
\begin{center}
\caption{Comparisons of inference time (in seconds per view) of different methods on the DTU dataset.}
\label{tabel:inference}
\begin{tabular}{c|ccccc}
\toprule[1.2pt]
\label{table:inference}
    &FVS & SVNVS &IBRNet&Ours-W&Ours-D\\ \hline
    Time (in seconds)&0.12&0.17&1.81&0.48&0.17
    \\
    
\bottomrule[1.2pt]
\end{tabular}
\end{center}
\vspace{-0.2cm}
\end{table}

\begin{figure}[thp]
\centering
\includegraphics[width=0.5\textwidth]{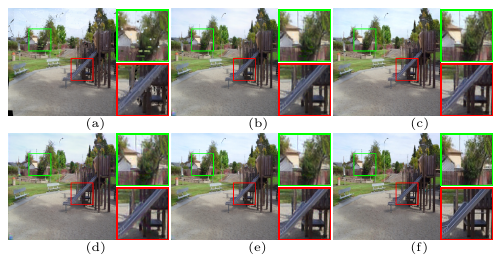}
\vspace{-0.4cm}
\caption{ 
Visual results of the ablation studies on network modules. (a)  Baseline. (b) Baseline + Inpainting. (c)  Baseline + Inpainting + Point Cloud Fusion. (d) Baseline + Inpainting + Refinement. (e) Complete model=  Baseline + Inpainting + Point Cloud Fusion + Refinement. (f) Ground truth. \color{cyan}{\faSearch~} Zoom in to see details.}
\label{figure:ablation}
\end{figure}

\begin{table}[tph]
\centering
\begin{center}
\caption{Quantitative results of the ablation studies. ``$\checkmark$"  (resp. ``$\times$") represents the corresponding module is used (resp. unused). 
} 
\vspace{-0.3cm}
\label{tabel:ablation}
\begin{tabular}{cccc|ccc}
\toprule[1.2pt]
\label{table:ablation}
    & Inpainting & Fusion & Refinement& \multicolumn{1}{c}{LPIPS$\downarrow$} & \multicolumn{1}{c}{PSNR$\uparrow$} & \multicolumn{1}{c}{SSIM$\uparrow$}  \\ \hline
	(1) & $\times$ &	$\times$ &$\times$ &	0.561&13.79&	0.362 
\\
	(2) & $\checkmark$&	$\times$ & $\times$ &0.157&22.97&	0.819	
\\
	(3) & $\checkmark$&	$\checkmark$&$\times$ & 0.138&24.18&	0.853	\\
	(4) & $\checkmark$&	$\times$&$\checkmark$ & 0.159&23.34&	0.836	\\
	(5) & $\checkmark$&	$\checkmark$&$\checkmark$ & \textbf{0.128}&	\textbf{24.77} &\textbf{0.869}\\
\bottomrule[1.2pt]
\end{tabular}
\end{center}
\vspace{-0.1cm}
\end{table}

\subsection{Ablation Study and Analysis}
\label{subsec:ablation}
\subsubsection{Network modules} We carried out comprehensive ablation studies on \textit{Tanks and Temples} with pre-processed depth maps as input to validate the effectiveness of the three key components of our framework, i.e., 
the image inpainting component, the point cloud fusion module, and the refinement component. We constructed a baseline by excluding these three components.  
We sequentially added each component to the base model until all three components were included to form the complete model. When removing the point cloud fusion module, we simply merged the sub-point clouds to generate a unified point cloud to enable the method.

\begin{figure}[thp]
\centering
\includegraphics[width=0.5\textwidth]{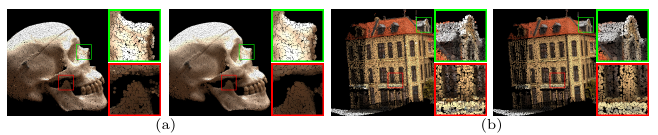}
\vspace{-0.4cm}
\caption{
Visual comparison of the 3D point clouds. The left images in (a) and (b) visualize the point clouds by directly projecting depth maps of input views into 3D space, and the right images show the point clouds learned by the proposed fusion method. }
\label{figure:pcs}
\end{figure}

Quantitatively, from Table~\ref{table:ablation},  
it can be seen that 
the reconstruction quality gradually improves with the inclusion of the three components, validating  
their effectiveness. 
Qualitatively, we showed the visual results of the ablation studies in Fig.~\ref{figure:ablation}. Compared with the baseline result, the proposed inpainting strategy 
can fill the holes and fill up the unknown area in the image boundary (see Fig.~\ref{figure:ablation} (\textcolor{red}{b})). 
With the proposed point cloud fusion module included, some local distortions and blurs effect are mitigated (see Fig.~\ref{figure:ablation} (\textcolor{red}{c})). 
The proposed refinement module promotes the synthesized image with more high-frequency details (see Fig.~\ref{figure:ablation} (\textcolor{red}{e})). 
The result of the point cloud fusion module exclusion experiment further demonstrates the effectiveness of our point fusion module (see Fig.~\ref{figure:ablation}(\textcolor{red}{d})). We also show the results of the directly projected point clouds and the unified point clouds learned by our fusion strategy in Fig.~\ref{figure:pcs}, where it can be seen that the unified point cloud learned by our fusion strategy has fewer noisy points and smoother edges.

\subsubsection{Point cloud fusion embeddings} We also validated the effectiveness of the embedded information contained in the point cloud fusion module, i.e., depth estimation quality and descriptor similarity. We conducted this experiment with our self-supervised depth module to demonstrate the effectiveness of the embedding term, depth estimation quality. As listed in Table~\ref{table:ablation_embedding}, it can be seen that each embedded information is helpful in improving the overall performance, verifying the rationality of our design. 

\begin{table*}[tph]
\centering
\begin{center}
\caption{Quantitative results of the ablation studies for the embedded information in point cloud fusion. ``$\checkmark$"  (resp. ``$\times$") represents the corresponding module is used (resp. unused). 
} \vspace{-0.2cm}
\begin{tabular}{ccc|ccc}
\toprule[1.2pt]
\label{table:ablation_embedding}
    \makecell{Relative position \& distance} & \makecell{Depth estimation quality} & \makecell{Descriptor similarity}& \multicolumn{1}{c}{LPIPS$\downarrow$} & \multicolumn{1}{c}{PSNR$\uparrow$} & \multicolumn{1}{c}{SSIM$\uparrow$}  \\ \hline
	$\checkmark$&	$\times$ & $\times$ &0.212&20.67&	0.724	
\\
	$\checkmark$&	$\checkmark$&$\times$ & 0.211&20.95&	0.733	\\
	$\checkmark$&	$\checkmark$&$\checkmark$ & \textbf{0.189}&	\textbf{21.28} &\textbf{0.751}\\
\bottomrule[1.2pt]
\end{tabular}
\vspace{-0.5cm}
\end{center}
\end{table*}

\subsubsection{Performance of our method under various numbers of input source views}
Due to the limited GPU memory, the proposed unified point cloud representation cannot directly process four sub-point clouds. Thus, we adopted a progressive fusion strategy, i.e., fusing three sub-point clouds as an intermediate unified point cloud further fused with the fourth sub-point cloud. Such a progressive manner can be straightforwardly extended to the case with more input views. As shown in Table~\ref{table:multiview}, the 3-input setting performs better than the 2-input because more information is available within the third input view. However, when given one more input view, there is no noticeable improvement. The main reason is that the camera motion in the dataset is very large, and the viewpoint of the fourth view is too far from the target view to provide useful information.

\begin{table*}[t]
\centering
\begin{center}
\caption{Quantitative results of our method with different view numbers on \textit{Tanks and Temples}.
} \vspace{-0.2cm}
\begin{tabular}{c|ccc|ccc|ccc|ccc}
\toprule[1.2pt]
\label{table:multiview}

  \# Input Views   & \multicolumn{3}{c|}{Train}    & \multicolumn{3}{c|}{Playground}            & \multicolumn{3}{c|}{M60}     & \multicolumn{3}{c}{Truck}    \\ \cline{2-13}
      & \multicolumn{1}{c}{LPIPS$\downarrow$} & \multicolumn{1}{c}{PSNR$\uparrow$} & \multicolumn{1}{c|}{SSIM$\uparrow$} & \multicolumn{1}{c}{LPIPS$\downarrow$} & \multicolumn{1}{c}{PSNR$\uparrow$} & \multicolumn{1}{c|}{SSIM$\uparrow$} & \multicolumn{1}{c}{LPIPS$\downarrow$} & \multicolumn{1}{c}{PSNR$\uparrow$} & \multicolumn{1}{c|}{SSIM$\uparrow$} & \multicolumn{1}{c}{LPIPS$\downarrow$} & \multicolumn{1}{c}{PSNR$\uparrow$} & \multicolumn{1}{c}{SSIM$\uparrow$} \\ \hline
2&0.145 &22.63&	0.835&\textbf{0.119}&27.50 &\textbf{0.887}&0.121 &24.78 &\textbf{0.897}&0.145&22.63&0.835\\
3 &\textbf{0.144}&23.13&0.828&0.125&\textbf{27.51}&\textbf{0.887}&0.119&24.94&0.891&0.110&\textbf{25.26}&0.868\\
4  &0.145 &\textbf{23.23}&	\textbf{0.831} &0.128 &27.49 &\textbf{0.887}& \textbf{0.115} &\textbf{25.02}& 0.893 &	\textbf{0.107} &25.23 &  \textbf{0.870}  \\
 \bottomrule[1.2pt]
 
\end{tabular}
\end{center}
\end{table*}

\begin{figure}[t]
\centering
\includegraphics[width=0.47\textwidth]{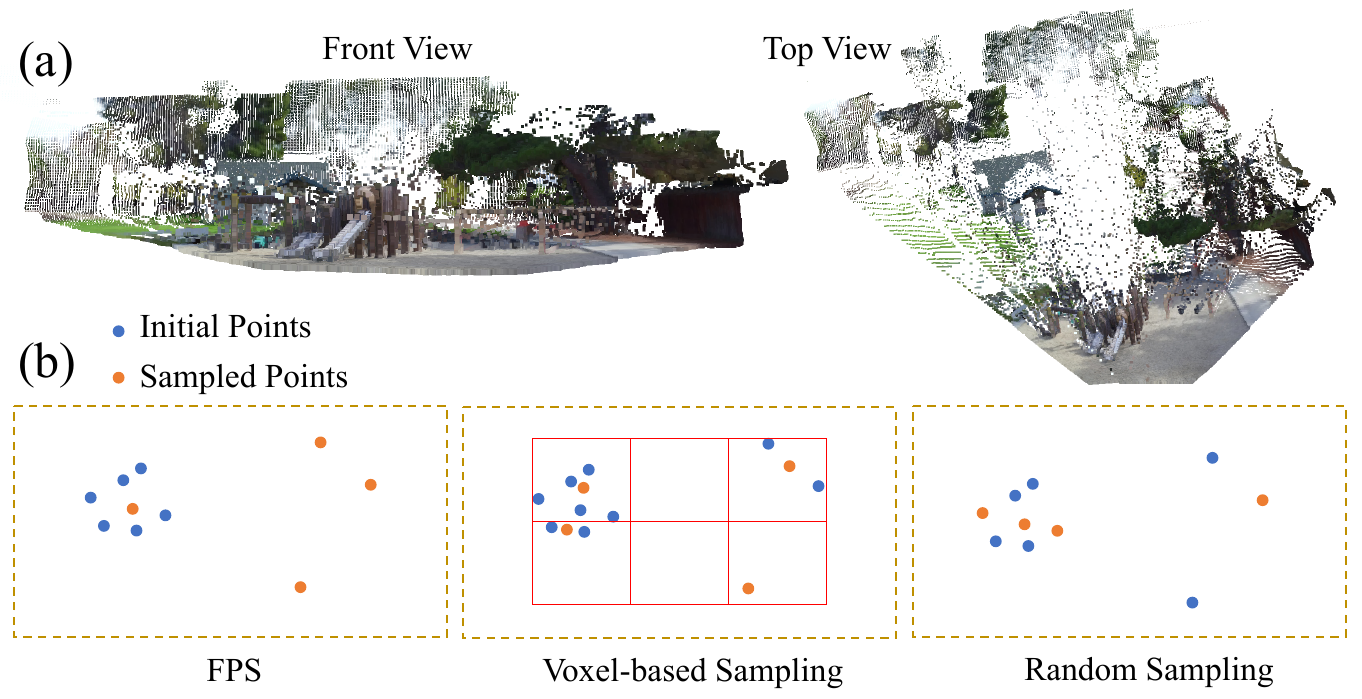} 
\vspace{-0.4cm}
\caption{(a) Projected point cloud in 3D space. (b)The operation of the three sampling methods on a set of ten points, in which seven points are scattered in a denser distribution on the left, and three points are scattered in a sparser distribution on the right. Specifically, FPS selects one point from the left distribution and all three points from the right distribution; Voxel-based sampling selects two points from the left distribution and two points from the right distribution; and Random sampling selects three points from the left distribution and one point from the right distribution.
}
\label{figure:sampling_p}
\end{figure}

\begin{figure}[t]
\centering
\includegraphics[width=0.5\textwidth]{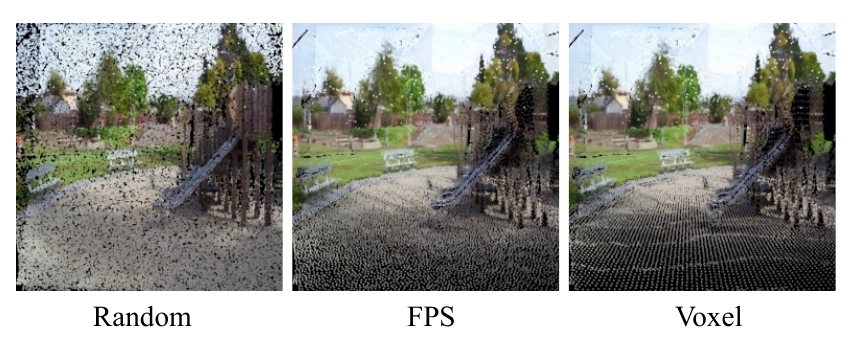} 
\vspace{-0.4cm}
\caption{Visual comparison of projected unified point cloud using different sampling strategies.
}
\label{figure:sampling}
\end{figure}
\subsubsection{Performance of our method under different sampling strategies}

We experimented with different sampling strategies to evaluate our method. We replaced random sampling used in \textit{Ours-D} with farthest point sampling (FPS) or voxel-based downsampling when performing inference. 
Given source views and their depth maps, we project the pixels of source views to a common 3D coordinate system using the pinhole camera model, producing sub-colored point clouds that are contained within frustum-shaped viewing volumes as shown in Fig.~\ref{figure:sampling_p} (\textcolor{red}{a}).
Since foreground objects of the scene typically account for a relatively larger image proportion and a smaller depth range, while background objects account for a smaller image proportion and a larger depth range, the points of foreground objects are denser than those of the background objects in the viewing frustum. Thus, the points in the union of these sub-point
clouds are also non-uniformly distributed in the 3D space, i.e., points of foreground objects are denser than those of background objects. Given such a non-uniform point cloud, we give a schematic diagram of their sampling methods in Fig.~\ref{figure:sampling_p} (\textcolor{red}{b}). Random sampling chooses each point in the union point cloud with the same probability, which results in sampling more points in foreground objects and fewer points in background objects. FPS selects a subset of points from a larger point cloud by iteratively selecting the point that is farthest from the previously selected points, resulting in sampled points being spread over the shape uniformly. When points of foreground objects are denser than those of background objects, FPS tends to sample a smaller percentage of points in the foreground objects while a larger percentage of points in the background objects,
leading to less preservation of foreground information than random sampling. Voxel sampling divides the 3D space into small voxels and only retains one point per voxel.
Since a typical voxel in foreground objects usually contains more points than the voxel in background objects, it leads to less preservation of foreground points after voxel sampling. 

We provided visual results of the unified point cloud projections from various sampling strategies in Fig.~\ref{figure:sampling}. It can be observed that the random sampling method preserves most foreground information from the original point cloud, while FPS and Voxel methods are prone to significant foreground information loss. Additionally, it is crucial that the sampled points cover the main geometry of the object. However, FPS and Voxel methods can compromise the structural integrity, especially when sampling a relatively small number of points, such as in the case of a broken wooden pile. We also show the quantitative results in Table~\ref{tabel:sampling}, where it is evident that random sampling outperforms the other techniques.

\subsubsection{Performance of our method under different hyperparameter settings}

\begin{figure}[t]
\centering
\includegraphics[width=0.30\textwidth]{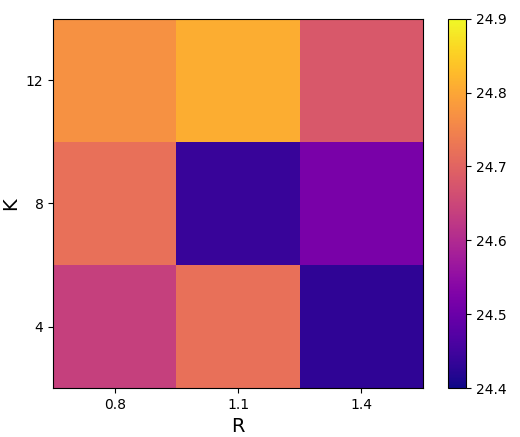} 
\vspace{-0.4cm}
\caption{The average PSNR on \textit{Tanks and Temples} on different settings of hyper-parameters $K$ and $R$.
}
\label{figure:kn}
\end{figure}

We investigated how the numbers of sampled points $M$ and neighboring point number $K$ affect the performance of  \textit{Ours-D} under the 2-input setting. Specifically, we tested different hyperparameter settings by changing the values of $K$ and $R$, where $M = R \cdot S$, $S = H \cdot W$. We compared the average PSNR across different settings on \textit{Tanks and Temples}, as shown in Fig.~\ref{figure:kn}. The small fluctuation in PSNR values across the different settings indicates the robustness of our model.
In our experiments, we adopt $K=12$ and $R=0.8$,  as a trade-off among the quality, the computation cost, and the memory cost. 
For the scenario with more input source views, we proportionally scale up the values of $K$ and $R$.

\begin{table*}[t]
\centering
\begin{center}
\caption{Quantitative results of the ablative study on the sampling strategy for obtaining anchor points on \textit{Tanks and Temples}.}
 \vspace{-0.2cm}
\label{tabel:sampling}
\resizebox{1\textwidth}{!}{
\begin{tabular}{c|ccc|ccc|ccc|ccc}
 \toprule[1.2pt] 
\label{table:inference}
 \# Sampling Strategy   & \multicolumn{3}{c|}{Train}    & \multicolumn{3}{c|}{Playground}            & \multicolumn{3}{c|}{M60}     & \multicolumn{3}{c}{Truck}    \\ \cline{2-13}
      & \multicolumn{1}{c}{LPIPS$\downarrow$} & \multicolumn{1}{c}{PSNR$\uparrow$} & \multicolumn{1}{c|}{SSIM$\uparrow$} & \multicolumn{1}{c}{LPIPS$\downarrow$} & \multicolumn{1}{c}{PSNR$\uparrow$} & \multicolumn{1}{c|}{SSIM$\uparrow$} & \multicolumn{1}{c}{LPIPS$\downarrow$} & \multicolumn{1}{c}{PSNR$\uparrow$} & \multicolumn{1}{c|}{SSIM$\uparrow$} & \multicolumn{1}{c}{LPIPS$\downarrow$} & \multicolumn{1}{c}{PSNR$\uparrow$} & \multicolumn{1}{c}{SSIM$\uparrow$} \\ \hline
      
Random&\textbf{0.145}&\textbf{22.63}&\textbf{0.835}
&\textbf{0.119}&\textbf{27.50}&\textbf{0.887}
&\textbf{0.121}&\textbf{24.78}&\textbf{0.897}
&\textbf{0.118}&\textbf{24.92}&\textbf{0.865}
\\
FPS
&0.146&21.96&0.830
&0.168&25.71&0.858
&0.140&23.78&0.882
&0.137&23.78&0.845
\\
Voxel&0.174&21.89&0.811
&0.219&24.18&0.822
&0.179&21.59&0.849
&0.195&22.29&0.806
\\
\bottomrule[1.2pt]
\end{tabular}
}
\end{center}
\vspace{-0.2cm}
\end{table*}

\subsubsection{Performance of our method under different orders of source views}
\label{subsubsec:order}
We conducted experiments to study how the order of source views affects Ours-D. Specifically, we examined the previously trained network under two scenarios: (i) only changing the order of warped source images in the concatenation operation in Eq.~\eqref{eq:refine}, and (ii) changing the order of input source images throughout the entire pipeline. For the 2-input setting, we swapped the order of the two input source views, and for the 3-input setting, we swapped the order of the first and third views. 
The changed order differs from the source view selection described in Section~\ref{sec:dataset} used for training. 
Table \ref{tabel:order} lists quantitative results, showing that modifying the order almost does not affect performance for the 2-input setting but results in a significant performance decrease for the 3-input setting. The possible reason is that, for the 2-input setting, both input views are usually near the target, whereas the added third view often has a more significant distant viewpoint from the target view in the 3-input setting. Thus, we should keep the source view selection consistent during training and inference to eliminate the order effect.

\begin{table*}[t]
\centering
\begin{center}
\caption{Quantitative results of the effect of the order of source views on the performance of \textit{Ours-D}.
}
\vspace{-0.3cm}
\label{tabel:order}
\resizebox{1\textwidth}{!}{
\begin{tabular}{cc|ccc|ccc|ccc|ccc}
\toprule[1.2pt]
    && \multicolumn{3}{c|}{Train}    & \multicolumn{3}{c|}{Playground}            & \multicolumn{3}{c|}{M60}     & \multicolumn{3}{c}{Truck}    \\ \cline{3-14}
      && \multicolumn{1}{c}{LPIPS$\downarrow$} & \multicolumn{1}{c}{PSNR$\uparrow$} & \multicolumn{1}{c|}{SSIM$\uparrow$} & \multicolumn{1}{c}{LPIPS$\downarrow$} & \multicolumn{1}{c}{PSNR$\uparrow$} & \multicolumn{1}{c|}{SSIM$\uparrow$} & \multicolumn{1}{c}{LPIPS$\downarrow$} & \multicolumn{1}{c}{PSNR$\uparrow$} & \multicolumn{1}{c|}{SSIM$\uparrow$} & \multicolumn{1}{c}{LPIPS$\downarrow$} & \multicolumn{1}{c}{PSNR$\uparrow$} & \multicolumn{1}{c}{SSIM$\uparrow$} \\ \hline

\multirow{3}{*}{2-input}&Ours-D &0.145 &22.63&	0.835&0.119&27.50 &0.887&0.121 &24.78 &0.897&0.145&22.63&0.835\\

&Change order of warped images
&0.146&22.64&0.832
&0.121&27.47&0.885
&0.119&24.90&0.900
&0.117&25.00&0.867
\\
&Change order of source images
&0.146&22.65&0.832
&0.121&27.44&0.885
&0.120&24.86&0.900
&0.116&25.01&0.867
\\\hline

\multirow{3}{*}{3-input}&Ours-D &0.144&23.13&0.828&0.125&27.51&0.887&0.119&24.94&0.891&0.110&25.26&0.868\\

&Change order of warped images
&0.157&22.60&0.812
&0.143&26.68&0.875
&0.134&24.29&0.878
&0.113&24.89&0.866
\\
&Change order of source images
&0.157&22.59&0.812
&0.143&26.65&0.874
&0.133&24.31&0.878
&0.113&24.89&0.866
\\
\toprule[1.2pt]
\end{tabular}
}
\end{center}
\vspace{-0.2cm}
\end{table*}

\section{Conclusion and Discussion}
\label{sec:conclusion}
We have presented a new learning-based paradigm for view synthesis, which learns a locally unified 3D point cloud representation from multiple source views. Precisely, we constructed the unified point cloud by adaptively fusing points at a local neighborhood defined on the union of the sub-point clouds projected from source views. Owing to the learning of the unified scene representation, as well as a 3D geometry-guided image restoration module to fill the holes and recover high-frequency details of the rendered novel view, our view synthesis paradigm reconstructs novel views with much higher quantitative and visual quality, compared with state-of-the-art methods.

In the future, the following directions could be considered for improving the proposed paradigm.
First, random sampling used in our point cloud fusion module to generate the base point clouds, may not be the optimal choice, and our paradigm will benefit from more advanced and efficient sampling methods. Second, efforts can be made to address the slight inconsistency between sub-point clouds, resulting from the progressive fusion strategy to balance the GPU memory cost when handling multiple source views. Third,  relative position or confidence map-aware algorithms could be investigated to mitigate the impact of the order of source views. Fourth, it is valuable to investigate inward-facing or long trajectory datasets to explore the potential of our framework in handling diverse scenarios. Finally, it would be interesting to investigate joint camera pose estimation and novel view synthesis for a self-contained paradigm.


%
%
\small
\bibliographystyle{IEEEtran}
\bibliography{egbib}
\end{document}